\documentclass[11pt]{article} 
\usepackage{url}
\usepackage{graphicx} 
\usepackage{epstopdf}
\usepackage[colorlinks, linkcolor=blue, anchorcolor=blue, citecolor=blue]{hyperref}
\usepackage[margin=1in]{geometry}
\usepackage[normalem]{ulem}
\usepackage[export]{adjustbox} 
\usepackage{mathtools, cuted}
\usepackage{enumerate}
\usepackage{enumitem}
\usepackage{microtype}
\usepackage{graphicx}
\usepackage{booktabs}
\usepackage{multirow} 
\usepackage{makecell}
\usepackage{amsmath}
\usepackage{amssymb}
\usepackage{mathtools}
\usepackage{amsthm}
\usepackage{algorithm}
\usepackage{algpseudocode} 
\usepackage{natbib}
\usepackage{listings}
\usepackage{subcaption}

\usepackage{kpfonts}
\DeclareMathAlphabet{\mathsf}{OT1}{cmss}{m}{n}

\SetMathAlphabet{\mathsf}{bold}{OT1}{cmss}{bx}{n}

\usepackage{amsmath}
\usepackage{amssymb}
\usepackage{mathtools}
\usepackage{amsthm}
\usepackage[symbol]{footmisc}

\usepackage{soul}

\usepackage{caption}

\usepackage[capitalize,noabbrev]{cleveref}

\theoremstyle{plain}

\theoremstyle{definition}

\theoremstyle{remark}

\usepackage[textsize=tiny]{todonotes}

\author{Zichong Li \thanks{Equal contribution}  $^1$, Liming Liu\footnotemark[1]  $^1$\footnote{Correspondence to \url{zli911@gatech.edu}, \url{lliu606@gatech.edu} and \url{tourzhao@gatech.edu}.} ,  Chen Liang$^2$, Weizhu Chen$^2$, Tuo Zhao$^1$ \\ $^1$Georgia Tech $^2$Microsoft}

\title{NorMuon: Making Muon more efficient and scalable}
\date{}

\begin{document}

\maketitle

\begin{abstract}
The choice of optimizer significantly impacts the training efficiency and computational costs of large language models (LLMs). Recently, the Muon optimizer has demonstrated promising results by orthogonalizing parameter updates, improving optimization geometry through better conditioning.
Despite Muon’s emergence as a candidate successor to Adam, the potential for jointly leveraging their strengths—has not been systematically explored.
In this work, we bridge this gap by proposing NorMuon (Neuron-wise Normalized Muon), an optimizer that synergistically combines orthogonalization with neuron-level adaptive learning rates. Our analysis reveals that while Muon effectively reduces condition numbers, the resulting updates exhibit highly non-uniform neuron norms, causing certain neurons to dominate the optimization process. NorMuon addresses this imbalance by maintaining second-order momentum statistics for each neuron and applying row-wise normalization after orthogonalization, ensuring balanced parameter utilization while preserving Muon's conditioning benefits. To enable practical deployment at scale, we develop an efficient distributed implementation under the FSDP2 framework that strategically distributes orthogonalization computations across devices.
Experiments across multiple model scales demonstrate that NorMuon consistently outperforms both Adam and Muon, achieving 21.74\% better training efficiency than Adam and 11.31\% improvement over Muon on 1.1B pretraining setting, while maintaining a comparable memory footprint to Muon. Our findings suggest that orthogonalization and adaptive learning rates are complementary rather than competing approaches, opening new avenues for optimizer design in large-scale deep learning. We open-source our implementation at \url{https://github.com/zichongli5/NorMuon.git}.
\end{abstract}


\section{Introduction}

Training efficiency remains a central challenge in scaling large language models (LLMs) \cite{dataefficienctllm,efficientllm}, where optimizer choice directly impacts convergence speed, computational requirements, and ultimately, the feasibility of training at scale \cite{muon,fantasticoptim}.
The community standard, Adam \citep{adam}, achieves robust performance through coordinate-wise preconditioning: dynamically adjusting learning rates for each parameter based on the second moment of its gradient history. While this per-coordinate adaptivity is computationally efficient and generally stable, it suffers from a fundamental limitation—it treats each parameter independently, ignoring the rich geometric structure and cross-coordinate dependencies inherent in neural network layers.

\begin{figure}[t]
\centering
\begin{subfigure}{0.45\textwidth}
    \centering
    \includegraphics[width=\linewidth]{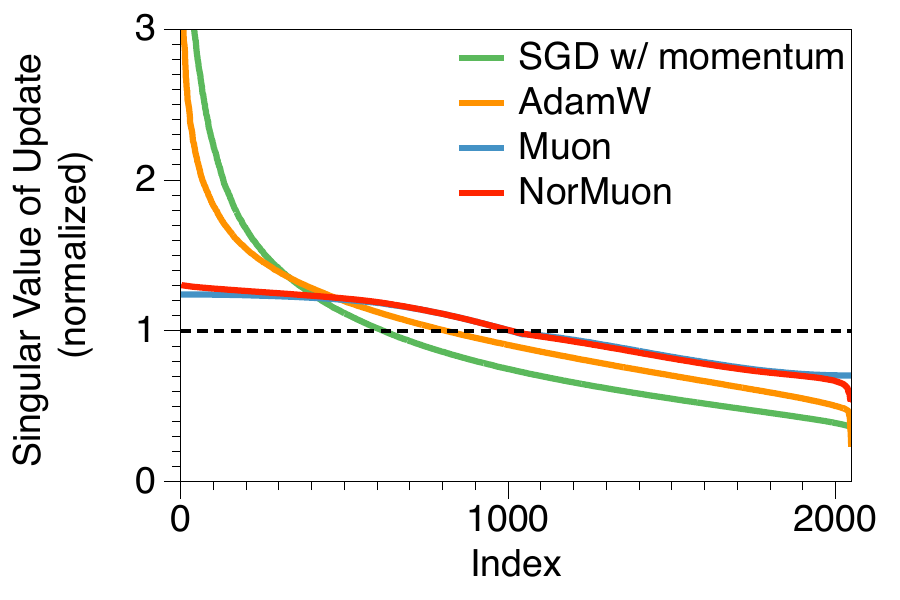}
    \caption{Singular value (sorted) of update directions}
    \label{fig:intro_svd}
\end{subfigure}
\hfill
\begin{subfigure}{0.45\textwidth}
    \centering
    \includegraphics[width=\linewidth]{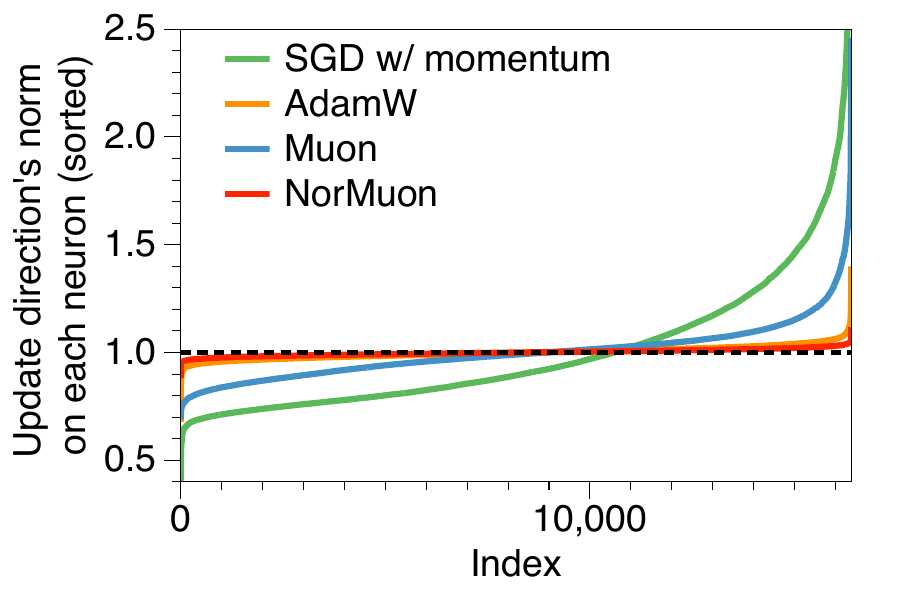}
    \caption{Per-neuron update norm}
    \label{fig:intro_norm}
\end{subfigure}
\caption{Analysis of optimization geometry during 1.1B model pretraining. We examine the up-projection matrix in the 8th layer's MLP at the middle checkpoint. (a) Singular value distribution reveals that raw momentum and AdamW's update exhibit high condition numbers. Muon's orthogonalization effectively eliminates this imbalance. (b) Despite Muon's improved conditioning, the $L^2$ norm of individual neuron updates still shows high variance. AdamW achieves much more uniform per-neuron norms. Our proposed NorMuon maintains Muon's low condition number while normalizing neuron contributions. We also include results for Adam-mini in Appendix \ref{appendix:intro}, which performs similarly with Adam.}
\label{fig:intro}
\end{figure}

Recent advances have sought to address this limitation through various approaches to capturing cross-coordinate structure. Adam-mini \citep{adammini} exploits the near-block-diagonal Hessian structure of neural networks by applying adaptive learning rates to parameter blocks (e.g. each neuron) rather than individual coordinates.
More ambitious second-order methods, such as Shampoo \citep{shampoo} and SOAP \citep{soap}, employ full-matrix preconditioning through singular value decomposition to capture curvature information and parameter interdependencies. However, these approaches incur substantial memory and communication overhead, while introducing hyperparameter sensitivity that limits their practical adoption at scale.

Recently, Muon \citep{muon} has emerged as a compelling middle ground, applying truncated Newton-Schulz iterations to approximate the orthogonal polar factor of momentum matrices. This approach yields matrix-wise orthogonalized updates that improve conditioning while maintaining modest computational overhead and approximately half the memory consumption of Adam, demonstrating promising results in LLM training.

These optimizers fundamentally differ in their preconditioning granularity and objectives. Adam and Adam-mini, when considered without exponential moving averages (EMAs), apply $L^2$ normalization at the per-coordinate and per-neuron levels respectively, adjusting learning rates while preserving update signs. In contrast, the idealized version of Shampoo and Muon operate at the per-matrix level, actively orthogonalizing parameter updates. 

The varied preconditioning strategies employed by these optimizers raise an important question: \ul{\textit{Are different forms of preconditioning inherently conflicting, or can they be combined in a way that yields complementary benefits?}}

To investigate this, we analyzed key properties of the update matrices from different optimizers during pretraining of a 1.1B-parameter Transformer model, examining both singular value distributions and per-neuron norms. As illustrated in Figure \ref{fig:intro_svd}, raw momentum accumulates updates with extremely high condition numbers, indicating that certain directions dominate while leaving other parameters underutilized. AdamW produces moderately more balanced singular values, though the improvement remains limited. In contrast, Muon's approximate orthogonalization successfully addresses this conditioning issue, yielding well-balanced singular values across the spectrum.
However, examining per-neuron update norms (Figure \ref{fig:intro_norm}) reveals a complementary perspective. AdamW demonstrates superior performance in reducing variance across per-neuron update norms compared to SGD momentum. Conversely, while Muon's orthogonalization effectively improves matrix-level conditioning, the per-neuron update norms exhibit high variance, with some neurons receiving disproportionately large updates relative to others.


This observation motivates our key insight: while Muon's orthogonalization effectively reduces the condition number of updates, the remaining high variance in neuron norms still creates an imbalanced learning dynamic, potentially leading to inefficient parameter usage. Drawing inspiration from Adam-mini's success \cite{adammini} with per-neuron adaptive learning rates, we propose to incorporate second-order momentum to normalize these disparate scales and ensure more balanced parameter updates. Our method, {\bf NorMuon}, augments Muon's orthogonalization with neuron-wise adaptive learning rates computed from accumulated second-order statistics. As demonstrated in our analysis, NorMuon yields updates with both low condition numbers (Figure \ref{fig:intro_svd}) and uniform neuron norms (Figure \ref{fig:intro_norm}), thereby combining the advantages of Muon and AdamW and achieving more balanced utilization of the network's representational capacity.

Beyond algorithmic innovation, the distributed implementation of orthogonalization-based optimizers remains relatively underexplored in the literature. To enable training at larger scales, we develop a distributed version of NorMuon compatible with the FSDP2 framework \citep{pytorch-fsdp2}. While previous work on distributed Muon \citep{muonscalable} was implemented using ZeRO-1 with Megatron-LM \citep{zero123, shoeybi2019megatron}, FSDP2 can offer greater flexibility and memory efficiency. However, direct adaptation of the previous distributed approach to FSDP2 would result in extensive replicated computation, as FSDP2 shards nearly all parameters across devices. Our implementation addresses this challenge by distributing orthogonalization computation across devices, eliminating redundant calculations while maintaining load balance. Furthermore, we leverage FSDP2's row-wise parameter sharding to enable efficient neuron-wise normalization without incurring additional communication overhead.

In summary, our contributions are threefold:

$\bullet$ We propose NorMuon, a simple and effective optimizer that combines Muon's orthogonalization with neuron-wise adaptive learning rates. NorMuon maintains uniform neuron norms to ensure balanced parameter utilization while preserving the low condition number achieved by Muon's orthogonalization.

$\bullet$ We develop an efficient distributed implementation under the FSDP2 framework. By carefully orchestrating sharded optimizer states, we gather updated momentum and distribute Muon orthogonalization computation uniformly across GPUs, achieving optimal memory efficiency with manageable communication and computational overhead.

$\bullet$ Through extensive experiments across multiple scales of LLM pretraining, we demonstrate that orthogonalization and blockwise adaptive learning rates are complementary rather than conflicting, with their combination yielding superior training dynamics compared to either approach in isolation.

\section{Related Works and Background}
\subsection{Related Works}

\hspace*{2em}\textbf{Adaptive Gradient Methods}. The introduction of per-parameter adaptive learning rates has been instrumental in training deep networks. Optimizers such as AdaGrad \citep{Adagrad}, RMSProp \citep{hinton2012rmsprop}, Adam \citep{adam} and AdamW \citep{adamw} use first- and second-moment estimates to adjust each weight’s step size individually. This coordinate-wise preconditioning improves stability and convergence in heterogeneous settings, and has become the de facto standard for LLM training. However, treating each weight independently ignores the underlying structure of neural network layers and incurs high memory overhead by storing two extra tensors per parameter. This memory cost motivated techniques like AdaFactor \citep{adafactor}, which factorizes the second-moment accumulator across rows and columns to reduce memory. Similarly, Adam-mini \citep{adammini} partitions parameters into blocks (e.g. each neuron’s weights) and assigns a single learning rate to each block, matching AdamW’s performance on different model sizes, while halving memory cost. GaLore \citep{galore} maintains momentum in a low-rank subspace derived from the SVD of gradients, although its effectiveness diminishes for long sequence lengths \citep{cosmos}.
Lion \citep{lion} applies a coordinate-wise signed update, abandoning second-order moment estimates to achieve memory savings.

\textbf{Second-order Methods}. In parallel, other optimizers capture the rich geometry of the loss surface by coupling updates across parameters.
K-FAC \citep{kfac} and its variants \citep{martens2018kronecker, gao2021trace} approximate curvature information beyond individual coordinates, capturing correlations across parameters. Shampoo \citep{shampoo} and its distributed variant \citep{distributedshampoo} employ Kronecker-factored preconditioners and have demonstrated strong performance in practice \citep{benchmarking}. More recently, SOAP \citep{soap} establishes a connection between Shampoo and Adafactor and further improves convergence performance. Despite these advances, Shampoo and SOAP incur substantial memory cost and computational overhead, which hinders their applicability at LLM scale.

\textbf{Orthogonal Update Methods}: Muon (Momentum Orthogonalized by Newton-Schulz, \cite{muon}) represents a breakthrough that leverages matrix geometry without the full cost of second-order methods. Muon performs an approximate polar decomposition (via Newton–Schulz iterations) on the momentum to extract its orthogonal component and also eliminates the need to store second-order momentum. Muon thus simultaneously improves both convergence and memory efficiency compared to Adam, demonstrating great potential for scaling up model pretraining \citep{muonscalable, practical}. We provide a detailed description of the Muon algorithm in Section \ref{intro:muon}.
More recently, Dion \citep{dion} extends the orthogonal update paradigm to be more communication- and compute-efficient in distributed settings. Dion applies a low-rank orthonormalization scheme via amortized power iteration instead of full Newton–Schulz, and decouples momentum buffers across devices to avoid full gradient synchronization.

\subsection{Background: Muon optimizer}
\label{intro:muon}

Muon \citep{muon} is an optimizer designed for the 2D weight matrices in neural network hidden layers. The key innovation lies in orthogonalizing the momentum before applying parameter updates, thereby improving the conditioning of the optimization trajectory.
Formally, at iteration $t$, given weight matrix $W_{t-1}$, learning rate $\eta_t$, and loss function $L$, Muon maintains a first-order momentum $M_t$ and computes updates as:
\begin{align}
M_t &= \mu\, M_{t-1} + \nabla L(W_{t-1}), \\
O_t &= \mathrm{NS5}(M_t), \\
W_t &= W_{t-1} - \eta_t\, O_t,
\end{align}where $M_0 = \mathbf{0}$ and $\mu$ is the momentum coefficient. The critical component is the orthogonalization operator $\mathrm{NS5}(\cdot)$, which aims to approximate the orthogonal projection of the momentum matrix:
\begin{align}
\text{Ortho}(M) = \arg \min_{O}\{\|O-M\|_{F}: O^{\top}O=I \text{ or } OO^{\top}=I\}.
\end{align}Muon approximates this orthogonalization through a fixed number of Newton-Schulz iterations. Starting with the Frobenius-normalized momentum $X_0 = M_t/\|M_t\|_F$, the algorithm performs $N$ iterations (typically $N=5$):
\begin{align}
X_k = a\,X_{k-1} + b\,(X_{k-1}X_{k-1}^{\top})\,X_{k-1} + c\,(X_{k-1}X_{k-1}^{\top})^{2}\,X_{k-1}, \quad k=1,\ldots,N,
\end{align}with the final orthogonalized update $O_t = X_N$. The coefficients $(a,b,c)$ are carefully chosen such that singular values of the update matrix converge toward unity. In practice, Muon is typically applied only to 2D weight matrices in hidden layers, while scalar parameters, bias vectors, embeddings, and unembedding layers continue to use standard optimizers such as Adam.

\section{Method}
In this section, we introduce NorMuon, where we aims to combine Muon's orthogonalization with block-wise adaptive learning rates based on the observation that the approximated orthogonalized updates can experience a high variance on update directions norm of each neuron.

\begin{algorithm}[ht]
\caption{NorMuon}
\label{algo:main}
\begin{algorithmic}[1]
\State \textbf{Input:} Initial weights $\mathbf{W}_0 \in \mathbb{R}^{m\times n}$, loss $L$, learning rate $\eta$, momentum parameters $(\beta_1, \beta_2)$, perturbation parameter $\varepsilon$, weight decay $\lambda$.\
\State Initialize $\mathbf{M}_0 \in\mathbb{R}^{m\times n} \leftarrow \mathbf{0}$, $\mathbf{v}_0 \in\mathbb{R}^{m}\leftarrow\mathbf{0}$
\For{$t=1,2,\ldots$}
\State $\mathbf{G}_t \leftarrow \nabla_{\mathbf{W}}L(\mathbf{W}_t)$
\State $\mathbf{M}_t \leftarrow \beta_1\mathbf{M}_{t-1} + (1-\beta_1)\mathbf{G}_t$
\State $\mathbf{O}_t \leftarrow \mathrm{NS5}(\mathbf{M}_t)$
\State $\mathbf{v}_t \leftarrow \beta_2\mathbf{v}_{t-1} + (1-\beta_2)\operatorname{mean}_{\text{cols}}(\mathbf{O}_t \odot \mathbf{O}_t)$
\State $\mathbf{V}_t \leftarrow \mathrm{ExpandRows}\left(\mathbf{v}_t\right)$ \hfill ($\mathbf{V}_t\in\mathbb{R}^{m\times n})$
\State $\widehat{\mathbf{O}}_t \leftarrow \mathbf{O}_t \oslash \left(\sqrt{\mathbf{V}}_t + \varepsilon\right)$ 
\State $\hat{\eta} = 0.2 \eta\sqrt{mn}/||\widehat{\mathbf{O}}_t||_F$
\State $\mathbf{W}_{t+1} \leftarrow \mathbf{W}_t - \eta\lambda\mathbf{W}_t - \hat{\eta}\widehat{\mathbf{O}}_t$
\EndFor
\end{algorithmic}
\end{algorithm}

\subsection{NorMuon}
\label{sec:method}
We present our update rule in Algorithm~\ref{algo:main}. The algorithm maintains two momentum states: the standard first-order momentum $\mathbf{M}_t \in \mathbb{R}^{m \times n}$ used by Muon (line 5), and an averaged second-order momentum $\mathbf{v}_t \in \mathbb{R}^{m}$ that tracks the squared magnitude of each neuron's update direction (lines 7).
Importantly, $\mathbf{v}_t$ requires minimal additional memory overhead, storing only $m$ scalars compared to the $m\times n$ first-order momentum.


At each iteration and given the gradient, we first follow the Muon's update rule that update the first-order momentum and apply Newton-Schulz iteration for orthogonalization (line 4-6), producing $\mathbf{O}_t$ with improved conditioning.
Rather than directly using this orthogonalized update, we compute row-wise statistics to capture the per-neuron update magnitudes. Specifically, we calculate the mean squared value across columns for each row of $\mathbf{O}_t$ (line 7). This statistic is accumulated into our averaged second-order momentum $\mathbf{v}_t$ using exponential moving average with decay rate $\beta_2$.
We then apply $\mathbf{v}_t$ for row-wise normalization (line 9).
This second-order momentum is similar to Adam-mini's \citep{adammini} block-wise reduced-dimensional statistics, where we treat each neuron (i.e. each row) as a block.
As illustrated in Figure 1, this normalization reduces the variance in update magnitudes across neurons while preserving the favorable conditioning properties.

We observe that after the row-normalization the resulting direction has a much larger norm. Hence, during the update, we add a learning rate scaling following \citep{muon} to keep a similar RMS norm to match Adam's RMS norm (line 10).

We would like to note that in the idealized case where the $\mathbf{O}_t$ is strictly orthogonalized (i.e., not approximated by NS5), the per-neuron norm would be strictly 1 for full-rank matrix with $m\leq n$. On these matrices, the neuron-wise normalization would not be beneficial. However, since orthogonalization is approximated in practice, we observe that this normalization remains necessary and helpful even for $m\leq n$ matrix (validated in Section \ref{sec:abla}).

\subsection{Distributed NorMuon}
As LLM training scales larger, distributed training becomes essential for both memory constraints and computational efficiency. We develop a distributed version of NorMuon compatible with the FSDP2 framework \citep{pytorch-fsdp2}, which employs ZeRO-3 style \citep{zero123} sharding to partition optimizer states, parameters, and gradients across multiple devices.

While coordinate-wise optimizers like Adam naturally extend to distributed settings, NorMuon presents unique challenges due to Muon's orthogonalization step, which requires access to complete momentum matrices. An existing distributed implementation of Muon \citep{muonscalable} gathers the full momentum on all devices and replicates the orthogonalization computation. We avoid such replicated costs by near-uniformly assign parameters to different devices.

Algorithm~\ref{alg:distributed} presents our distributed implementation. The key modifications from Algorithm~\ref{algo:main} are:

\noindent $\bullet$ \textbf{Efficient Orthogonalization Distribution} (line 5-9): Rather than having all devices compute orthogonalization for all parameters, we first sort the parameter list by matrix size (line 2) to ensure uniform work assignment, where $\text{Numel}(\cdot)$ counts the number of elements in each matrix. We then assign each parameter tensor to a specific device using a round-robin scheme. Only the assigned device gathers the full momentum matrix via all-gather communication and performs the Newton-Schulz orthogonalization (lines 5-8), before scattering the result back to all devices (line 9). This approach eliminates redundant computation while maintaining load balance across devices.

\noindent $\bullet$ \textbf{Shard-Local Row Normalization} (lines 10-12): A key advantage of our design is that the row-wise normalization operates entirely on local shards without additional communication. This is possible because FSDP2 employs row-wise sharding, ensuring that each device holds complete rows of the weight matrix. The computation of row statistics and normalization thus proceed independently.

\begin{algorithm}[ht]
\caption{Distributed NorMuon: one iteration}
\label{alg:distributed}
\begin{algorithmic}[1]
\State \textbf{Input:} Sharded 2D weights $\{\mathbf{W}_{\text{shard}}^{(i)}\}_{i=0,...,N}$, Sharded gradient $\{\mathbf{G}_{\text{shard}}^{(i)}\}_{i=0,...,N}$, learning rate $\eta$, momentum parameters $(\beta_1, \beta_2)$, perturbation parameter $\varepsilon$, weight decay $\lambda$. We omit the initialization of optimizer states for simplicity.\
\State $\{\mathbf{W}_{\text{shard}}^{(i)}\}_{i=0,...,N} \leftarrow \text{Sort}(\{\mathbf{W}_{\text{shard}}^{(i)}\}_{i=0,...,N}, \text{key}=\text{Numel}(\cdot))$
\For{$i=0,1,\ldots, N$}
\State $\mathbf{M}_{\text{shard}}^{(i)} \leftarrow \beta_1\mathbf{M}_{\text{shard}}^{(i)} + (1-\beta_1)\mathbf{G}_{\text{shard}}^{(i)}$
\If {$i \bmod  \text{world size} == \text{current rank}$}
\State $\mathbf{M}^{(i)} \leftarrow \text{Gather}(\mathbf{M}_{\text{shard}}^{(i)})$
\State $\mathbf{O}^{(i)} \leftarrow \mathrm{NS5}(\mathbf{M}^{(i)})$
\EndIf
\State $\mathbf{O}_{\text{shard}}^{(i)} \leftarrow \text{Scatter}(\mathbf{O}^{(i)})$
\State $\mathbf{v}_{\text{shard}}^{(i)} \leftarrow \beta_2\mathbf{v}_{\text{shard}}^{(i)} + (1-\beta_2)\operatorname{mean}_{\text{cols}}(\mathbf{O}_{\text{shard}}^{(i)} \odot \mathbf{O}_{\text{shard}}^{(i)})$
\State $\mathbf{V}_{\text{shard}}^{(i)} \leftarrow \mathrm{ExpandRows}\left(\mathbf{v}_{\text{shard}}^{(i)}\right)$
\State $\widehat{\mathbf{O}}_{\text{shard}}^{(i)} \leftarrow \mathbf{O}_{\text{shard}}^{(i)} \oslash \left(\sqrt{\mathbf{V}}_{\text{shard}}^{(i)} + \varepsilon\right)$ 
\State $\hat{\eta} = 0.2 \eta\sqrt{\text{Numel}(\widehat{\mathbf{O}}_{\text{shard}}^{(i)})}/||\widehat{\mathbf{O}}_{\text{shard}}^{(i)}||_F$
\State $\mathbf{W}_{\text{shard}}^{(i)} \leftarrow \mathbf{W}_{\text{shard}}^{(i)} - \eta\lambda\mathbf{W}_{\text{shard}}^{(i)} - \hat{\eta}\widehat{\mathbf{O}}_{\text{shard}}^{(i)}$
\EndFor
\end{algorithmic}
\end{algorithm}

\subsection{Overhead Analysis}

\hspace*{2em}\textbf{Memory Overhead}. NorMuon maintains Muon's memory efficiency. For a weight matrix $W \in \mathbb{R}^{m\times n}$, the memory consumption of optimizer states for each optimizer is: (1) Adam: $2mn$ (first and second-order momentum); (2) Muon: $mn$ (first-order momentum only); (3) NorMuon: $m(n+1)$ (first-order momentum + per-neuron second-order statistics).
The additional memory overhead of NorMuon compared to Muon is negligible (1/n factor), while remaining approximately 50\% more memory-efficient than Adam.

\textbf{Communication Overhead}. NorMuon introduces moderate additional communication compared to standard FSDP training. Under FP32 training with AdamW, the per-parameter communication cost is:
4 bytes (forward all-gather) + 4 bytes (backward all-gather) + 4 bytes (gradient reduce-scatter) = 12 bytes.
With NS5 iteration computed in BF16 precision, NorMuon requires:
12 bytes (standard FSDP communication) + 2 bytes (momentum gather, BF16) + 2 bytes (update scatter, BF16) = 16 bytes.

This represents a 33\% increase in communication volume. When parameters use BF16, the relative overhead increases to 50\%. However, this communication can be overlapped with orthogonalization computation to minimize latency impact. In our experiments (Section \ref{sec:efficiency}), we demonstrate that the per-iteration latency of NorMuon is only 3\% higher than AdamW, while achieving significantly better convergence efficiency.

\section{Experiments}
\label{sec:exp}
In this section, we conduct pretraining experiments across four different model scales to validate the effectiveness of NorMuon: 124M, 350M, 1.1B, and 5.4B parameters. 
For the larger models (1.1B and 5.4B), we adopt the experimental setup from previous work on architecture scaling \citep{phi4miniflash}, with results and configurations presented in Section \ref{sec:slimpajama}. For the smaller models (124M and 350M), we follow the experimental setting of Modded-NanoGPT \citep{nanogpt}, with results and settings provided in Section \ref{sec:nanogpt}.
We include extensive ablation studies that justify our design choices, along with detailed efficiency analyses (Section \ref{sec:abla} and \ref{sec:efficiency}).

\subsection{Experiments on 1.1B and 5.4B Models}
\label{sec:slimpajama}

\subsubsection{Setup}
\hspace*{2em}\textbf{Models.} We follow a simple linear rule from prior works \citep{scalinglaw, phi4miniflash} for scaling the architectural shape. Specifically, for a model with depth $d$ layers, we configure the architecture as follows: hidden dimension $\alpha d$, number of attention query heads $d$, number of attention key-value heads $d/4$, and MLP intermediate dimension $4\alpha d$, where $\alpha = 128$. The $\alpha$ and ratios are derived relative to Llama-3-8B \citep{llama3}. Our 1.1B and 5.4B parameter models correspond to depths of $d = 16$ and $d = 28$ layers, respectively.

\textbf{Dataset.} We conduct pretraining on the SlimPajama dataset \citep{slimpajama} and train our models on 50B tokens.

\textbf{Hyperparameters.} We employ Depth-$\mu$p \citep{depthmup} to scale the learning rate inversely proportional to $\sqrt{d}$ based on model depth. The base learning rate is set to $4 \times 10^{-4}$ with a base model depth of 16. The learning rate schedule consists of a 1B token warmup phase followed by linear decay to 0. 
We apply 0.1 weight decay for 2D parameters in hidden layers and zero weight decay for others to enhance training stability \citep{phi4miniflash}. The batch size is fixed at 2M tokens with a sequence length of 4096 tokens. 
For optimization, we use the following configurations: Adam optimizer with $(\beta_1, \beta_2) = (0.9, 0.95)$, Muon optimizer with $\beta_1 = 0.95$ following \cite{muon}, and our proposed NorMuon optimizer with $(\beta_1, \beta_2) = (0.95, 0.95)$.

\textbf{Baselines.} We compare NorMuon against three established optimizers: AdamW \citep{adamw}, the standard adaptive optimizer with decoupled weight decay; Muon \citep{muon}, which applies orthogonalization to update directions; and Dion \citep{dion}, a scalable orthogonalization-based method that uses low-rank power iteration. For orthogonalization-based optimizers, we apply orthogonalization to the concatenated QKV matrix rather than separately for a fair comparison, as we observe no performance improvement with separate application for Muon.

\subsubsection{Main results}

\begin{figure}[ht]
\centering
\begin{subfigure}{0.45\textwidth}
    \centering
    \includegraphics[width=\linewidth]{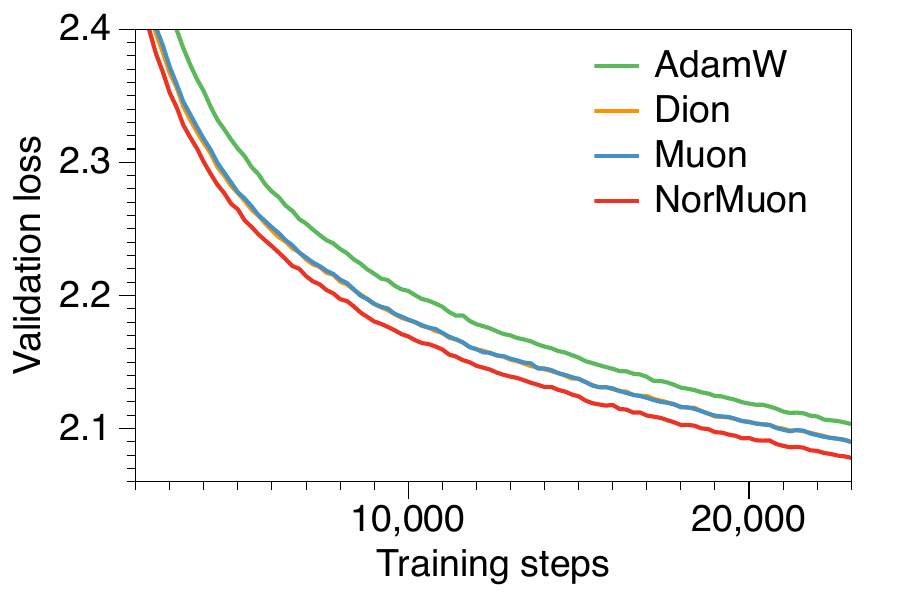}
    \caption{Pretraining results of 1.1B model.}
\end{subfigure}
\hfill
\begin{subfigure}{0.45\textwidth}
    \centering
    \includegraphics[width=\linewidth]{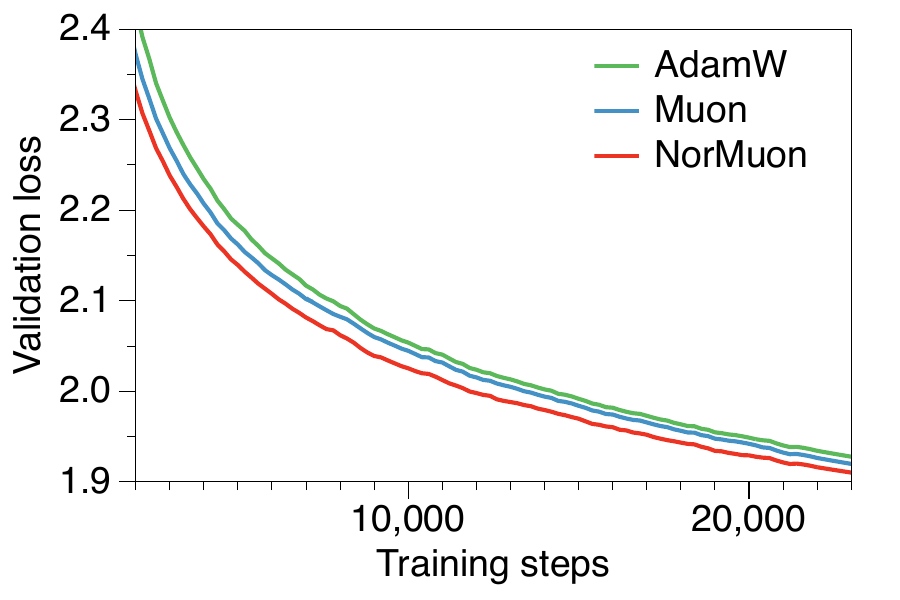}
    \caption{Pretraining results of 5.4B model.}
\end{subfigure}
\caption{Comparison of different optimizers on pretraining on 1.1B (a) and 5.4B (b) Transformers. NorMuon outperforms other baselines by notable margin.}
\label{fig:val_loss_1_5B}
\end{figure}

Figure \ref{fig:val_loss_1_5B} presents the validation loss curve across different model scales. NorMuon demonstrates consistent and substantial improvements over all baseline optimizers. While orthogonalization-based optimizers (Muon and Dion) already outperform AdamW, NorMuon amplifies this advantage through the integration of our proposed neuron-wise adaptive learning rate.

To quantify the performance gains, Table \ref{tab:efficiency-gain_1_5B} reports the percentage reduction in training steps required for each optimizer to achieve the same final validation loss as Adam. NorMuon achieves the best efficiency gains of 21.74\% and 13.91\% for the 1.1B and 5.4B models, respectively.

\begin{table}[ht]
\centering
\caption{Efficiency Gain over Adam. Calculated as percentage reduction in training steps required to reach the same final loss achieved by Adam. Dion's performance on 5.4B model is excluded due to resource constraints and similar performance with Muon on 1.1B scale.}
\label{tab:efficiency-gain_1_5B}
\resizebox{0.47\linewidth}{!}{
\begin{tabular}{lcc}
\toprule
\textbf{Optimizer} & \textbf{1.1B Model (\%)} & \textbf{5.4B Model (\%)} \\
\midrule
Muon     & 10.43 & 6.08  \\
Dion     & 10.43 & --    \\
NorMuon  & 21.74 & 13.91 \\
\bottomrule
\end{tabular}
}
\end{table}

\subsubsection{Ablation studies.} 
\label{sec:abla}

To validate our design choices in NorMuon, we conduct ablation studies that examine three key aspects: the granularity of adaptive learning rates and the positioning of normalization relative to orthogonalization, and the impact of applying normalization universally versus selectively based on matrix dimensions. The results on pretraining 1.1B model are presented in Figure \ref{fig:1B_abla}.

\textbf{Adaptive Learning Rate Granularity}. We compare our neuron-wise adaptive approach against coordinate-wise adaptation in the ``Muon+Adam" baseline, which applies Adam's coordinate-wise normalization after Muon's orthogonalization. This approach is similarly explored in concurrent work \citep{adamuon}, though we removed the sign stabilization component as we found it performs better without it. While Muon+Adam demonstrates improvements over vanilla Muon, it underperforms NorMuon across the training trajectory. Importantly, this approach requires maintaining full second-order momentum of orthogonalized updates, doubling the memory overhead (Section \ref{sec:efficiency}).

\textbf{Normalization Positioning}. We further investigate whether the positioning of adaptive normalization matters by testing ``NorMuon (Front)", which applies neuron-wise adaptive learning rates before orthogonalization rather than after. This variant still improves upon Muon, but slightly underperforms NorMuon.

\textbf{Universal vs. Selective Normalization}: To test whether normalization is beneficial to $m\leq n$ matrices as discussed in Section \ref{sec:method}, we evaluate "NorMuon ($m>n$ only)", which applies normalization only to $m>n$ matrices.
We can see that this selective approach underperforms the full NorMuon, demonstrating that applying normalization to those with $m \leq n$ is helpful.

\begin{minipage}{0.47\textwidth}
  \begin{figure}[H]
    \centering
    \includegraphics[width=\linewidth]{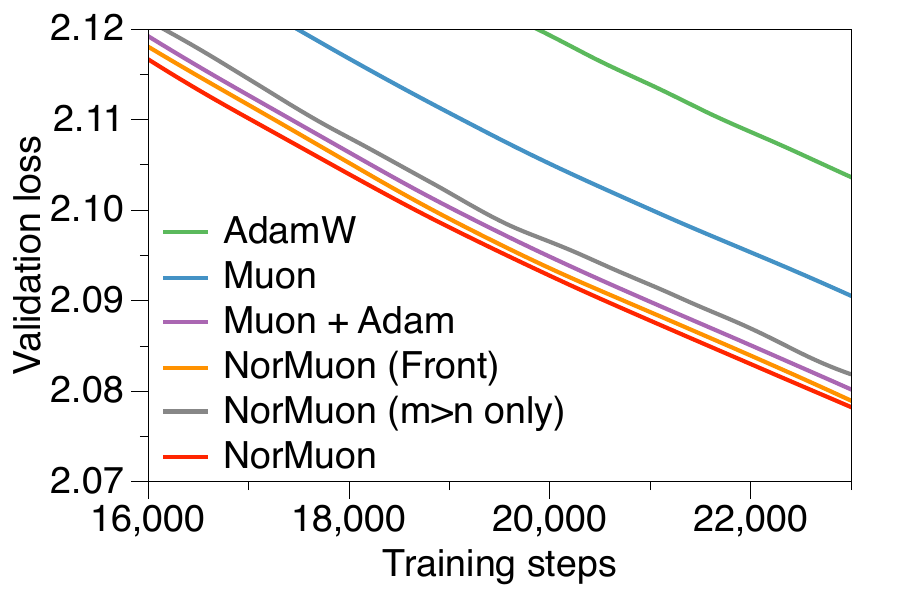}
    \caption{Ablation studies of NorMuon on 1.1B pretraining experiments.}
    \label{fig:1B_abla}
  \end{figure}
\end{minipage}\hfill
\begin{minipage}{0.47\textwidth}
  \begin{figure}[H]
    \centering
    \includegraphics[width=\linewidth]{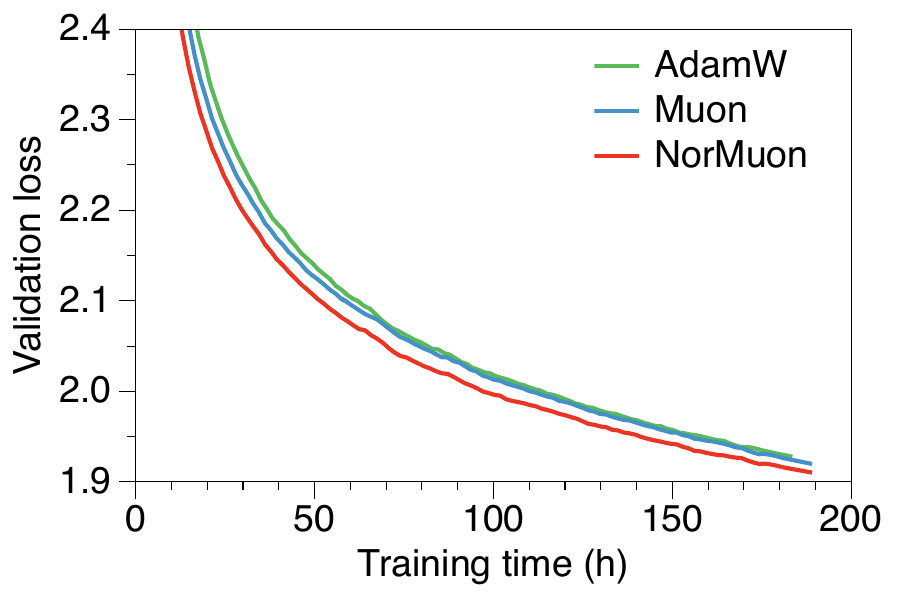}
    \caption{Validation loss vs wallclock time on 5.4B pretraining experiments.}
    \label{fig:wall_clock_5B}
  \end{figure}
\end{minipage}

\subsubsection{Computational and Memory Overhead Analysis}
\label{sec:efficiency}
In this section, we demonstrate that the overheads incur by NorMuon are manageable and do not diminish the practical benefits.

\textbf{Wall-Clock Performance}. Figure \ref{fig:wall_clock_5B} presents the validation loss as a function of wall-clock training time. Despite the additional computation required for orthogonalization and neuron-wise normalization, NorMuon maintains substantial performance advantages over AdamW.

\textbf{Memory and Computational Overhead}. Table \ref{tab:overhead_analysis} provides a breakdown of the computational and memory requirements for each optimizer. NorMuon achieves comparable memory efficiency to Muon with nearly a 50\% reduction compared to AdamW or Muon + Adam.
In terms of computational cost, NorMuon introduces only a 2.9\% increase in training step time compared to AdamW. The neuron-wise norm computation adds minimal cost relative to orthogonalization operations. 
Our efficient orthogonalization distribution strategy across GPUs proves crucial for maintaining low overhead. Without proper work distribution, the optimizer step time increases to approximately 2.7×.
Notably, this strategy can and has been applied to Muon for fair comparison.

\begin{table}[ht]
\centering
\caption{Computational and memory overhead comparison for different optimizers when training a 5.4B parameter model. Training step time includes forward pass, backward pass, and optimizer step. Percentages indicate relative increase compared to AdamW baseline.}
\label{tab:overhead_analysis}
\begin{tabular}{lcccc}
\toprule
\textbf{Optimizer} & \makecell{Memory cost of \\optimizer states (GB)} &  \makecell{Optimizer step\\ time (s)} & \makecell{Training step\\ time (s)} \\
\midrule
AdamW     & 40.56 & 0.02 & 28.73  \\
Muon    & 21.14 &  0.83 & 29.56 (2.8\%$\uparrow$)\\
Muon + Adam   & 40.56 &  0.85 & 29.58 (3.0\%$\uparrow$)\\
\midrule
NorMuon  & 21.19 & 0.84 & 29.57  (2.9\%$\uparrow$)\\
w/o orthogonalization distribution  & - & 2.29 & 31.04  (8.1\%$\uparrow$)\\
\bottomrule
\end{tabular}
\end{table}



\begin{figure}[ht]
\centering
\begin{subfigure}{0.45\textwidth}
    \centering
    \includegraphics[width=\linewidth]{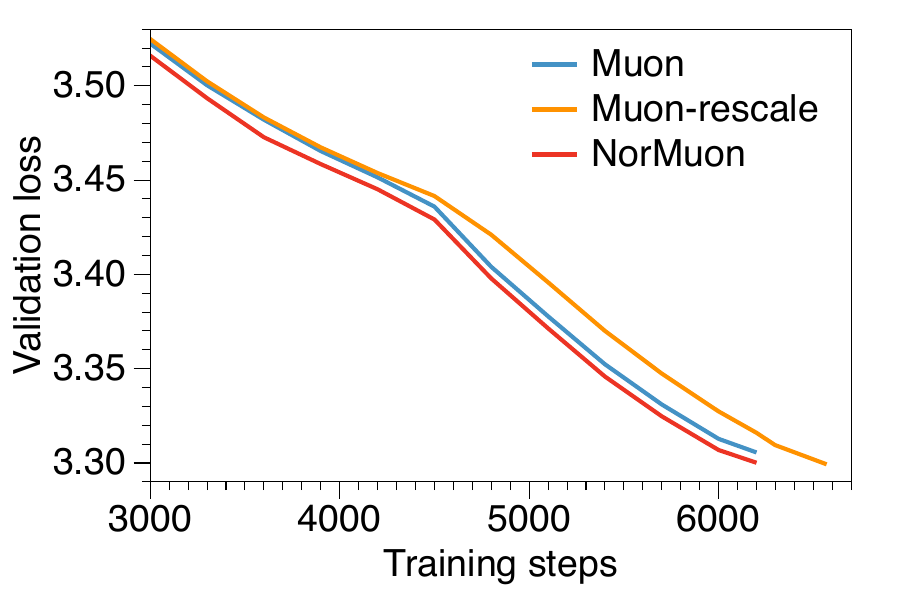}
    \caption{Pretraining results of 124M model.}
\end{subfigure}
\hfill
\begin{subfigure}{0.45\textwidth}
    \centering
    \includegraphics[width=\linewidth]{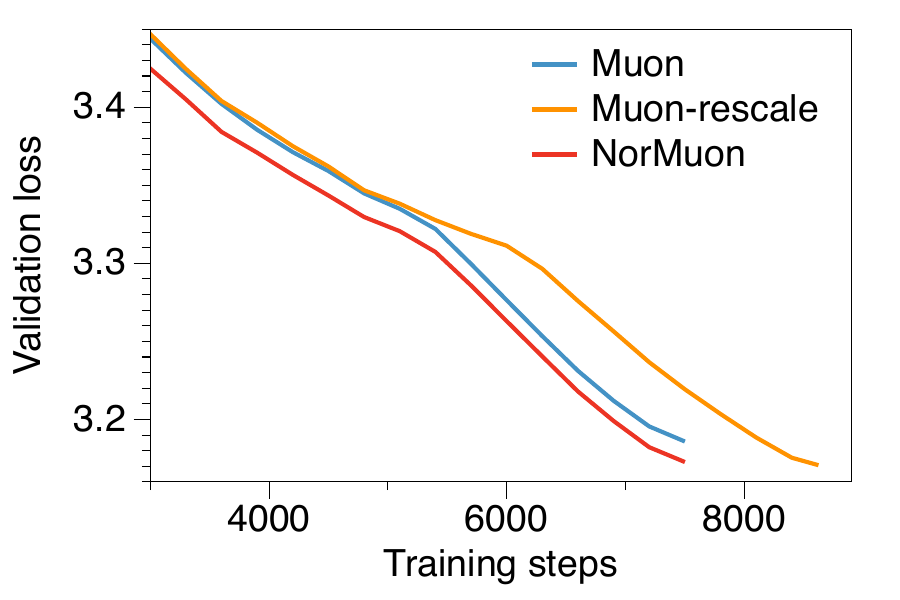}
    \caption{Pretraining results of 350M model.}
\end{subfigure}
\caption{Comparison of Muon and NorMuon on pretraining 124M (a) and 350M (b) Modded-NanoGPT on FineWeb. NorMuon outperforms Muon by notable margin.}
\label{fig:modded-nanogpt}
\end{figure}

\subsection{Experiments on Modded-NanoGPT}
\label{sec:nanogpt}
To further verify the advantages of NorMuon over Muon, we conduct experiments using Muon's original experimental setting on Modded-NanoGPT \citep{nanogpt}. Detailed experimental configurations and more ablation results are provided in Appendix~\ref{appendix:modded} and \ref{appendix:abla_modded}, respectively.

\textbf{Main Results}. Figure~\ref{fig:modded-nanogpt} presents the comparison between NorMuon and Muon on 124M and 350M parameter models. NorMuon consistently outperforms Muon across both model sizes. Since Muon's improvements over Adam have been extensively demonstrated in \citet{muon}, we omit those baseline comparisons here.

To quantify the computational benefits of NorMuon, we conduct an additional analysis where Muon is trained with the same learning rate schedule but for a longer total number of iterations, until it reaches the same validation loss as NorMuon (denoted as ``Muon-rescale" in Figure~\ref{fig:modded-nanogpt}). We observe that on the 124M model, Muon requires 6\% more iterations than NorMuon to achieve the same validation loss. On the 350M model, this efficiency gap increases substantially to 15\%, demonstrating the broad advantage of NorMuon over Muon.

\section{Conclusion}

In this work, we introduced \textbf{NorMuon}, a simple yet effective optimizer that integrates Muon’s orthogonalization with neuron-wise adaptive learning rates.
To make NorMuon practical for large-scale training, we developed an efficient distributed implementation under the FSDP2 framework, carefully orchestrating momentum gathering and orthogonalization to eliminate redundant computation and communication overhead.
Our experiment results shows notable improvement over Muon, demonstrating that orthogonalization and adaptive scaling need not be mutually exclusive, but can be complementary, when combined, lead to superior optimization dynamics.

\section*{Acknowledgments}
We thank Liliang Ren for valuable discussions and feedback regarding the code and its implementation.

\bibliography{reference}
\bibliographystyle{ims}


\newpage
\appendix
\section{Additional Experiments}
\subsection{Adam-mini's results on optimization geometry analysis}
\label{appendix:intro}
\begin{figure}[h]
\centering
\begin{subfigure}{0.45\textwidth}
    \centering
    \includegraphics[width=\linewidth]{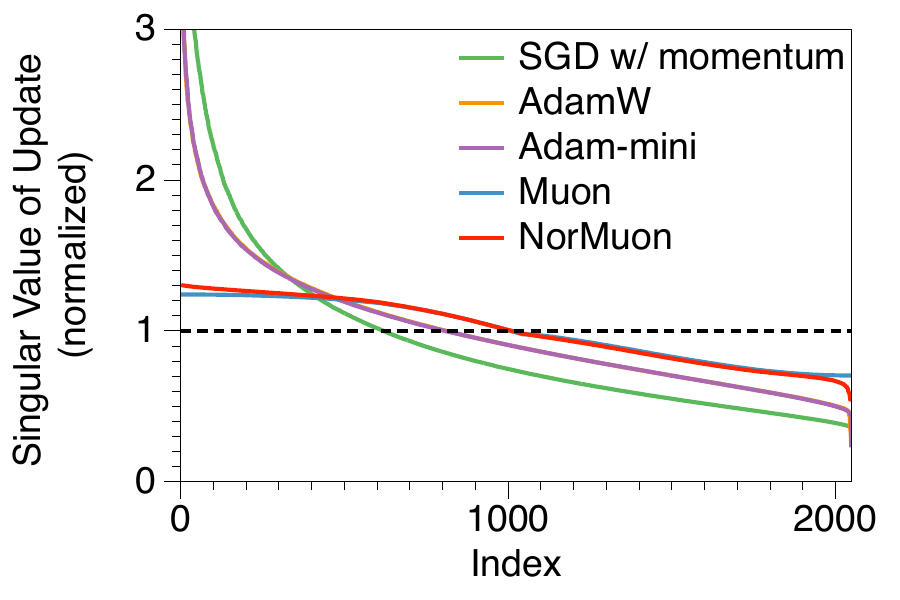}
    \caption{Singular value (sorted) of update directions}
\end{subfigure}
\hfill
\begin{subfigure}{0.45\textwidth}
    \centering
    \includegraphics[width=\linewidth]{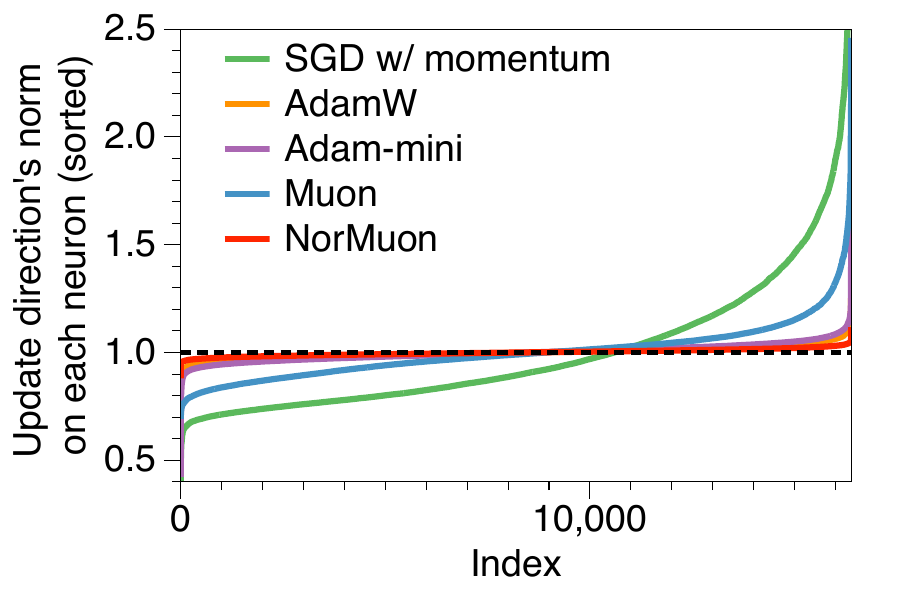}
    \caption{Per-neuron update norm}
\end{subfigure}
\caption{Analysis of optimization geometry during 1.1B model pretraining for different optimizers.}
\end{figure}

\subsection{Ablation Experiments on Modded-NanoGPT}
\label{appendix:abla_modded}
\begin{figure}[ht]
\centering
\includegraphics[width=0.6\linewidth]{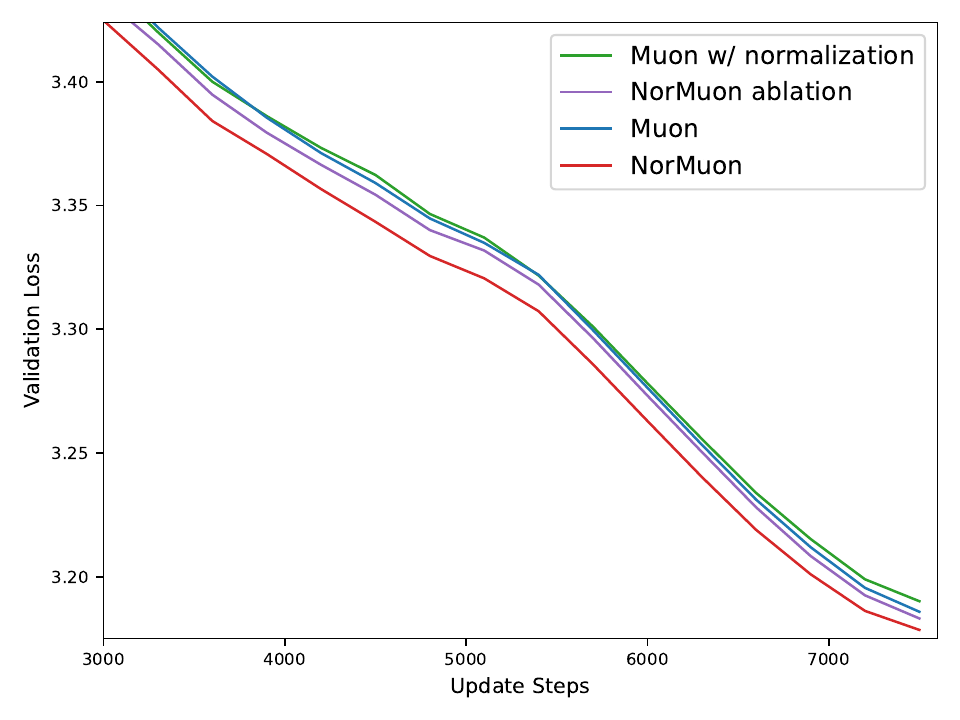}
\caption{Ablation studies on pretraining 350M model.}
\label{fig:modded-nanogpt-ablation}
\end{figure}

To further verify the effectiveness of NorMuon, we conducted several ablation experiments under the setting of training a 350M Modded-NanoGPT on FineWeb, and show the results in Figure \ref{fig:modded-nanogpt-ablation}:

{\bf (1)} Standard NorMuon, denoted as "NorMuon" in Figure \ref{fig:modded-nanogpt-ablation}.

{\bf (2)} Standard Muon used in original setting of Modded-NanoGPT \citep{muon}, denoted as "Muon" in Figure \ref{fig:modded-nanogpt-ablation}.

{\bf (3)} Applying normalization directly to Muon's update such that the update is strictly $\sqrt{m \times n}$, denoted as "Muon w/ normalization" in Figure \ref{fig:modded-nanogpt-ablation}.

{\bf (4)} applying NorMuon only to weight matrices with $m > n$, while using the normalized muon mentioned in (3) for all other weight matrices, denoted as "NorMuon ablation" in Figure \ref{fig:modded-nanogpt-ablation}.

We can see that although Muon with normalization performs slightly better than Muon in the early stages, it is eventually surpassed by Muon, indicating that the effectiveness of NorMuon cannot be attributed to normalization. Furthermore, since weight matrices with $m > n$ correspond only to the MLP up-projection matrices, which constitute only a small portion of the model, applying NorMuon only to this subset of parameters greatly diminishes the effect of NorMuon, resulting in only a marginal improvement over Muon.

\section{Implementation Details}
For experiments involving 1.1B and 5.4B parameter models, we conducted training on 2 nodes, each equipped with 8 A100 GPUs (80GB) connected via NVLink for optimized inter-GPU communication. Training duration was approximately 2 days for the 1.1B model and 7 days for the 5.4B model.

All experiments using Modded-NanoGPT, including both 124M and 350M parameter models, were performed on a single node with 8 A100 GPUs.

\section{Experiment Setup of Modded-NanoGPT}
\label{appendix:modded}

We strictly follow the experimental setup of Muon \citep{muon}, with details below:

\textbf{Models}: The model architecture is consistent with GPT-2 \citep{gpt2}, with 124M and 350M parameter configurations obtained by adjusting width and depth.

\textbf{Dataset}: We train all models on the FineWeb dataset \citep{fineweb}. The 124M model is trained on approximately 3.2B tokens, while the 350M model uses approximately 4B tokens.

\textbf{Hyperparameters}: Since Muon has already performed extensive hyperparameter tuning in this setting \citep{muon}, we adopt their optimized configurations except for $\beta_1$, which we slightly tune. We use a batch size of 512, sequence length of 1024, and the Warmup-Stable-Decay (WSD) learning rate schedule. Training iterations are set to 6,200 for the 124M model and 7,500 for the 350M model.

For the 124M model, Adam's parameters uses a learning rate of $3.6 \times 10^{-3}$ with momentum parameters $(\beta_1, \beta_2) = (0.9, 0.95)$. For Muon and NorMuon, we set the learning rate to $3.6 \times 10^{-4}$ and conduct a grid search over $\beta_1 \in \{0.9, 0.95\}$, reporting the best result. For NorMuon, $\beta_2$ is set to 0.95.

For the 350M model, Adam employs differentiated learning rates: $0.3$ for the embedding layer and $3 \times 10^{-3}$ for the output layer, with momentum parameters $(\beta_1, \beta_2) = (0.8, 0.95)$. For hidden layers, Muon and NorMuon use a learning rate of $7.5 \times 10^{-4}$, with $\beta_1$ selected from $\{0.9, 0.95\}$ based on validation performance.

\end{document}